\documentclass[review]{elsarticle}

\usepackage{graphicx}

\usepackage{amssymb}
\usepackage{adjustbox}

\usepackage{lineno}
\modulolinenumbers[5]

\usepackage{amsmath}
\usepackage{amsfonts}   
\usepackage{amssymb}   %

\usepackage{amsthm}   

\usepackage{textcomp}

\usepackage{algorithm} 
\usepackage{algorithmic}

\usepackage[caption=false,font=footnotesize]{subfig}

\usepackage{url}

\usepackage{xcolor} 
\usepackage{enumitem}

\usepackage[utf8]{inputenc}

\usepackage{multirow}
\usepackage{svg}
\usepackage{adjustbox}
\usepackage{amsmath}

\usepackage{booktabs}

\theoremstyle{definition} 

\journal{Expert Systems with Applications}

\begin{document}

\begin{frontmatter}

\title{Towards a Data Privacy-Predictive Performance Trade-off}

\author[label1]{Tânia Carvalho\corref{cor1}}
\ead{tania.carvalho@fc.up.pt}

\author[label1,label2]{Nuno Moniz}
\ead{nmmoniz@inesctec.pt}

\author[label3]{Pedro Faria}
\ead{pbfaria@tekprivacy.com}

\author[label1]{Luís Antunes}
\ead{lfa@fc.up.pt}

\address[label1]{Department of Computer Science, Faculty of Computer Science, University of Porto, Rua do Campo Alegre, 1021, 4169-007 Porto, Portugal}
\address[label2]{INESC TEC, Rua Dr. Roberto Frias 4200-465 Porto, Portugal}
\address[label3]{Faculty of Computer Science, University of Porto, Rua do Campo Alegre, 1021, 4169-007 Porto, Portugal}

\cortext[cor1]{Corresponding author}

\begin{abstract}

Machine learning is increasingly used in the most diverse applications and domains, whether in healthcare, to predict pathologies, or in the financial sector to detect fraud. One of the linchpins for efficiency and accuracy in machine learning is data utility. However, when it contains personal information, full access may be restricted due to laws and regulations aiming to protect individuals' privacy. Therefore, data owners must ensure that any data shared guarantees such privacy. Removal or transformation of private information (de-identification) are among the most common techniques. Intuitively, one can anticipate that reducing detail or distorting information would result in losses for model predictive performance. However, previous work concerning classification tasks using de-identified data generally demonstrates that predictive performance can be preserved in specific applications. In this paper, we aim to evaluate the existence of a trade-off between data privacy and predictive performance in classification tasks. We leverage a large set of privacy-preserving techniques and learning algorithms to provide an assessment of re-identification ability and the impact of transformed variants on predictive performance. Unlike previous literature, we confirm that the higher the level of privacy (lower re-identification risk), the higher the impact on predictive performance, pointing towards clear evidence of a trade-off.

\end{abstract}


\end{frontmatter}


\section{Introduction}\label{sec:intro}

Privacy rights is a topic that has been discussed in the computing field for over 50 years~\citep{hoffman1969computers}. At that time, awareness of individuals' privacy led to the beginning of suggested legal and administrative safeguards appropriate to the computerised world. Since then, privacy has disputed several legal problems around the world. Although there is great utility and many advantages in technological evolution, there is an increase in information sharing where all data is re-used and analysed on an unprecedented scale. 
Citizens are increasingly concerned about what companies and institutions do with the data they provide and how it is handled. To protect the data subjects' privacy, a world-wide legislative effort is observed, of which the General Data Protection Regulation\footnote{Regulation (EU) 2016/679 of the European Parliament and of the Council of 27 April 2016} (GDPR) is a prime example, unifying data privacy laws across Europe. Entities must determine appropriate technical and organisational measures to process personal data to comply with this regulation. To determine which measures are appropriate, it is essential to distinguish between data formats. A very recent survey on heterogeneous data was introduced by~\cite{cunha2021survey}. The authors propose a privacy taxonomy that establishes a relation between different types of data and suitable privacy-preserving strategies for the characteristics of those data types. In our study, we focus on structured data sets.

Data is used in the development and update of machine learning tools to, for example, ensure useful product recommendations, timely fraud detection, and to facilitate accurate medical diagnoses. 
However, the constraints of privacy-protecting legislation may have a necessary impact on the performance of such applications. Legislation enforces that prior to data sharing ventures, organisations must ensure that all private information concerning individuals is de-identified, i.e., to detach individuals from their personal data.
De-identification procedures typically reduce available information and data granularity, with possible consequences for the predictive performance of machine learning tools. Such limitation describes the challenge that third parties face when developing machine learning solutions using de-identified data.

The tension between the level of privacy and the predictive accuracy has been confirmed in many studies~\citep{fung2010privacy, kifer2011no}. Nevertheless, related work concerning the usage of de-identified data for classification tasks, in general, demonstrates a diversity of methods that allow obtaining a desired level of privacy without compromising predictive performance. However, the conclusions are limited, as no extensive experiments have been carried out to corroborate the results, i.e., previous works generally do not have a wide variety of data sets. Furthermore, despite the concern in guaranteeing the individuals' privacy, it is observed that most of the proposed works do not present the privacy risks that transformed data may still have. Most importantly, the evidence of a trade-off between re-identification ability and predictive performance has not yet been clearly provided.

In this paper, our main goal is to assess the existence of a trade-off between identity disclosure risk and predictive performance. To achieve such a goal, we implement all steps of a de-identification process for privacy-preserving data publishing. This includes leveraging combinations of several privacy-preserving techniques. Consequently,  we assess re-identification risk and predictive performance to investigate the impact of such techniques.
We briefly summarise our contributions as follows:
\begin{itemize}
    \item We provide an extensive experimental study comprising of 62 data sets, including the impact of five privacy-preserving individual techniques and their combination in predictive performance;
    \item We analyse the existence of a potential trade-off between the re-identification ability and predictive performance;

\end{itemize}

The remainder of the paper is organised as follows: Section~\ref{sec:litreview} provides a literature review and discussion of related work concerning data de-identification and machine learning, namely classification tasks. The problem definition is formulated in Section~\ref{sec:privperf}. The extensive experimental evaluation and the obtained results are provided in Section~\ref{sec:expstudy}. Section~\ref{sec:discussion} presents a discussion of the previous results. Finally, conclusions are presented in Section~\ref{sec:conclusions}. 

\section{Literature Review}\label{sec:litreview}

The need to protect personal information has led to the proposal of several privacy-preserving techniques (PPT) in related literature, in which the main goal is to allow multiple data use to its fullest potential without compromising individuals' privacy. As a result, privacy-preserving data publishing emerged to provide methods capable of protecting confidential data so they can be released to other parties safely while maintaining its utility.

Before applying the PPT, it is fundamental to know the different attributes in order to assign the most suitable technique. Accordingly, the attributes commonly take the following form~\citep{domingo2016database}:
\begin{description}
    \item \textbf{Identifiers}: attributes such as name and social security number 
    that directly identifies an individual.

    \item \textbf{Quasi-identifiers (QI)}: 
    the combination of these attributes may lead to an identification, e.g. date of birth, gender, geographical location and profession.

    \item \textbf{Sensitive}: highly critical attributes, usually protected by law and regulations, e.g. religion, sexual orientation, health information, political opinion and ethnic group.

    \item \textbf{Non-sensitive}: other attributes that do not fall in the above categories.
\end{description}

Based on the previous attributes description, data controllers must apply robust PPT to obtain de-identified data sets. In such transformation, we should consider three common types of privacy threats~\citep{wp29}: \textit{i)} singling out (often known as re-identification or identity disclosure), which means that an intruder can isolate records that identify an individual in the data set; \textit{ii)} linkability, which denotes the ability to connect or correlate two or more records of an individual or a group of individuals and \textit{iii)} inference, which corresponds to the possibility to deduce the value of an attribute based on the values of other attributes.

An intruder, also known as an adversary, is an individual who possesses skills, resources and motivation to re-identify the data subjects or deduce new information about them in the de-identified data set. 
De-identification effectiveness is determined by how likely an intruder will be able to disclose personal information on any individual. A traditional method to measure the ability on re-identification is record linkage~\citep{fellegi69}, used to link two records. 

To avoid re-identification attacks, data must be transformed in such a way that privacy risks are considerably reduced. However, transformation involves reducing available information or its distortion, which would expect to result in a reduction of data utility~\citep{brickell2008cost}. Therefore, the de-identification process must also consider an analysis of data utility. In classification tasks, utility references predictive performance -- the closer the evaluation results obtained between the original and the transformed data, the more utility is preserved. In other words, data utility is measured by the quality of the protected data set. The typical metrics used to assess data mining and machine learning predictive performance include Precision and Recall~\citep{Kent_prec_acc}, Accuracy, 
F-score~\citep{Rijsbergen_fmeasure}, AUC (Area Under the ROC Curve)~\citep{weng2008new} and G-Mean~\citep{kubat1998machine}. For additional detail,~\citet{fletcher2015measuring} describe several data utility measures.

\begin{figure}[ht!]
   \centering
   \scriptsize
   \includegraphics[width=0.8\linewidth]{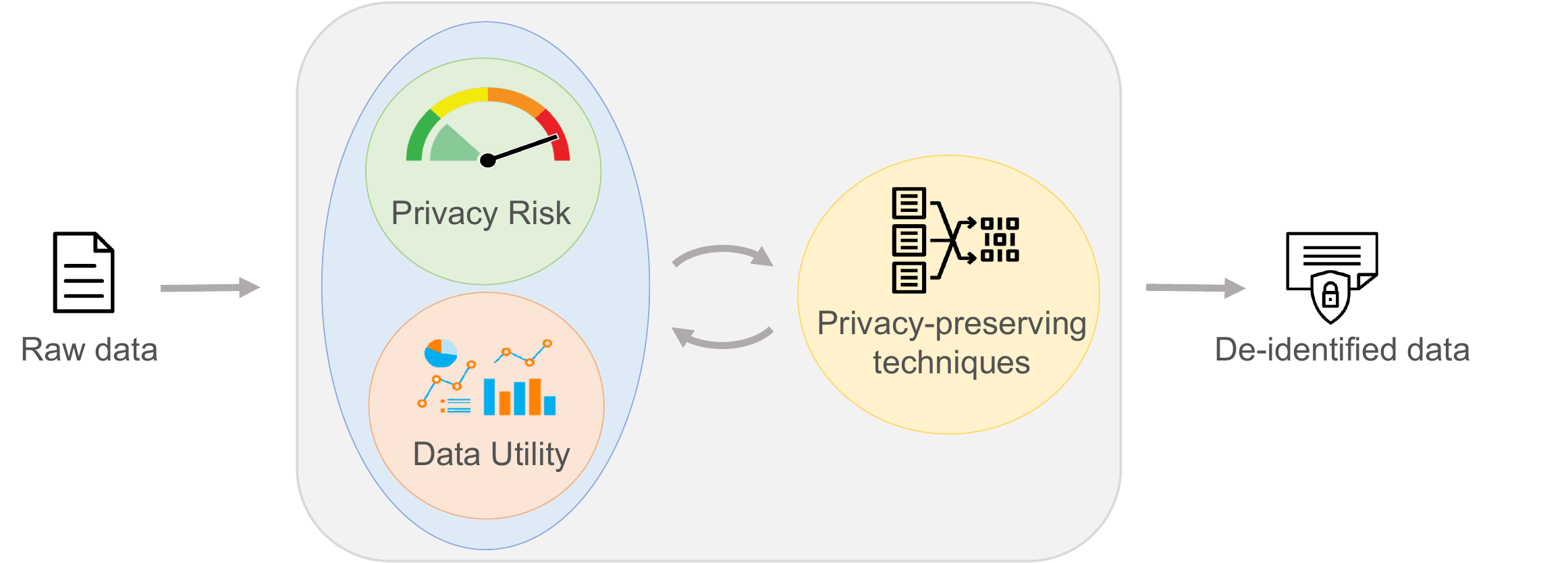}
 \caption{De-identification process in privacy-preserving data publishing.}
 \label{fig:flux}
\end{figure}

A general overview of the main steps in the de-identification process applied is presented in Figure~\ref{fig:flux}. The first step is to assess the raw disclosure risk and data utility. The choice of PPT is then applied based on the need for data protection determined by the disclosure risk, structure of data, and attributes. After data transformation, the disclosure risk and data utility is re-assessed. If the compromise between the two measures is not met, the last steps are repeated; otherwise, the data is protected and can be released.

Over the years, several PPT to limit disclosure risk were proposed~\citep{domingo2008survey, torra2017masking, murthy2019comparative}. The most popular approaches to obtain a protected data set are classified as: \textit{i))} non-perturbative, which reduces the amount of information in the original data, either suppressing some data or reducing the level of detail and; \textit{ii))} perturbative, which unlike non-perturbative, distorts the original data. 

Global re-coding, top-and-bottom coding and suppression are some examples of non-perturbative category. The idea of global re-coding is to group  the values into broader categories, and in the case of continuous attributes, discretisation may be applied. Top-and-bottom coding~\citep{koberg2016measuring} is similar to global re-coding, but only applied to ordinal categorical attributes or continuous attributes: values above or below a certain threshold are re-coded. Both techniques are considered a subgroup of generalisation. Regarding suppression, values are replaced by a missing value (NA) or special character (*, ?). This can be applied in three different levels: cell, record or attribute. \cite{samarati1998generalizing} were the first to propose generalisation and suppression to achieve protected data.

Noise~\citep{Brand2002, kim2003multiplicative} and rounding~\citep{senavirathne2019rounding} techniques belongs to the perturbative category. With noise application, the original data set values are changed by adding or subtracting some random value. The rounding technique replaces the original values of selected attributes with rounded values, which are integer multiples of a rounding base. 

The data controller can enforce a specific measure for data privacy using a single PPT or a combination of techniques.~\citet{prasser2020flexible} provide an overview of data privacy measures. The most known privacy measure, and the first being proposed, was $k$-anonymity by~\citet{samarati2001protecting}. 
This measure indicates how many $k$ occurrences appears in the data set concerning a certain combination of QI values. A record is unique when $k=1$. For better protection, $k$ must be great than 1. In addition, no single sensitive attribute can be dominant in an equivalence class. A very similar measure is $l$-diversity~\citep{machanavajjhala2007diversity}, showing that intruders with more background knowledge can deduce sensitive information about individuals even without re-identifying them. This measure, indicates how many $l$-``well represented'' values are in each sensitive attribute for each equivalence class. 
$\epsilon$-Differential privacy, introduced by~\citet{dwork2008differential}, has received considerable attention, recently. A randomised query function satisfies $\epsilon$-differential privacy if the result of differential-private analysis of an individual's information is the same whether or not that individual's information is included in the input for the analysis. Extensions of this method have been proposed to differentially private data release~\citep{zhu2017differentially}. Generally, $\epsilon$-differential privacy is a method used to apply Laplace noise addition. Hence, we refer to $\epsilon$-differential privacy as a noise-based method.

Several researchers have analysed the impact of data set de-identification in terms of predictive performance, the main topic of our contribution. The following section provides a description of such work, along with a discussion on existing limitations and open questions.

\subsection{Related Work}\label{subsec:relwork}

\citet{iyengar2002transforming} was the first to evaluate the impact of de-identified data in the context of classification tasks. The author transforms the original data to satisfy privacy constraints using both generalisation and suppression techniques. Then, the author evaluates the $k$-anonymous data sets according to the performance of models built using such sets. Other studies have followed Iyengar's example with the exception of the methodology used for generalisation and suppression application~\citep{wang2004bottom, fung2005top}. Such studies prove that de-identification level increase leads to proportional degradation of predictive performance. However, it is still possible to protect individuals' privacy while maintaining predictive performance with both techniques. 
While the previous works use more than one data set in their experiments, many studies were also presented~\citep{buratovic2012effects, sridhar2012evaluating, paranthaman2013performance, de2017privacy}, but using a single data set to investigate the effect of de-identified data in classification tasks. The conclusions are similar to the previous ones.

\citet{lefevre2006workload} presented a suite of de-identifications methods to generate de-identified data based on target workloads, which consists of several data mining tasks. Their procedure attempts to minimise the amount of generalisations required to reach a certain $k$-anonymity, while considering the data utility in the generalisation step. The authors proved the effectiveness of such proposal by achieving better results in classification than previous approaches. However,~\citet{li2011information}, achieved better results in classification than~\citet{lefevre2006workload} proposal. Their method determines the level of generalisation by the attributes distribution instead of a specific $k$ correspondent to the privacy requirement. Regarding suppression, it is applied to remove locally detailed information.

Among the several studies and proposals using both generalisation and suppression, none of those highlight identity disclosure risk. Recently,~\citet{vanichayavisalsakul2018evaluation}, transform the original data to respect different privacy measures, using these PPT. The authors also presented the re-identification risk for each transformation and showed a case with a lower level of re-identification risk and higher accuracy. The conclusion indicates that there is no significant difference in predictive performance between the original data and the de-identified data.

Regarding the noise application,~\citet{mivule2013comparative} propose a comparative classification error gauge. Noise is applied to the original data and the noisy data is passed through classifiers, where the respective models' predictive performance is obtained. If the level of errors in predictive performance is greater than a specified threshold, the noise and classifier parameters are re-adjusted; otherwise the best utility was achieved. The procedure is repeated until the desired predictive performance is reached. The results suggest a trade-off at approximately 20\% misclassification, but the experiments was carry out with a single data set.

Noise could also be added through a specific mechanism to reach the well-known privacy level measure, $\epsilon$-differential privacy~\citep{dwork2008differential}. A few works propose a methodology that applies random noise with a Laplace mechanism \citep{mivule2012towards, zorarpaci2020privacy}.
Their experiments show that the level of added noise does have an impact on the predictive performance. 
Nonetheless, the best result was obtained with higher $\epsilon$ values. Such a result is expected, given that the higher $\epsilon$ is, less noise will exist~\citep{lee2011much}. However, higher $\epsilon$ is not recommended~\citet{dwork2011firm}. Besides that, recent study shows that setting a certain $\epsilon$ does not provide the confidentiality offered by this method~\citep{muralidhar2020epsilon}; therefore, theoretical guarantees of $\epsilon$-differential privacy are challenged. 

In recent work,~\citet{tania2021compromise} apply three PPT (generalisation, suppression and noise) along with possible combinations between them. Conclusions point towards a noticeable impact of such techniques in predictive performance. Besides that, for each transformed data set, the authors conduct a re-identification risk evaluation through record linkage. 

Table~\ref{tab:refs} summarises the main bibliographic references of related literature. We denote ``No'' in the impact of private data on classification results when the authors consider a small degradation of those results and ``Yes'', otherwise.

\begin{table}[ht!]
\begin{center}
    \scriptsize
    \begin{adjustbox}{max width=0.85\textwidth}
\begin{tabular}{@{}llll@{}}
\toprule
\textbf{\begin{tabular}[c]{@{}l@{}}Privacy-preserving\\ techniques\end{tabular}}              & \textbf{\begin{tabular}[c]{@{}l@{}}Re-identification \\ risk \end{tabular}}  & \textbf{\begin{tabular}[c]{@{}l@{}}Performance \\ impact\end{tabular}} & \textbf{References} \\ \midrule
\multirow{2}{*}{\begin{tabular}[c]{@{}l@{}}Generalisation\\ and suppression\end{tabular}} &
No    & No     &  \begin{tabular}[c]{@{}p{5.5cm}@{}}
\cite{iyengar2002transforming, wang2004bottom, fung2005top, lefevre2006workload, li2011information, buratovic2012effects, sridhar2012evaluating, paranthaman2013performance, de2017privacy}     \end{tabular}    \\ \cmidrule(l){2-4} 
  & Yes    & No       & \cite{vanichayavisalsakul2018evaluation}           \\ \midrule
  
\multirow{2}{*}{\begin{tabular}[c]{@{}l@{}}Noise\end{tabular}}  & No   &  No    &  \cite{mivule2013comparative}           \\ 
\cmidrule(l){2-4} 
& No & Yes & \cite{mivule2012towards, zorarpaci2020privacy} \\ \midrule

\begin{tabular}[c]{@{}l@{}}Generalisation, \\ suppression and noise\end{tabular}
     & Yes      & Yes     & \cite{tania2021compromise}            \\
     \bottomrule
\end{tabular}
\end{adjustbox}
\end{center}
\caption{Main references of related work.}
\label{tab:refs}
\end{table}

\subsection{Summary and Novelty}\label{subsec:litreview_sum}

Since the introduction of the privacy measure $k-$anonymity, several proposals have immediately emerged to protect data subjects privacy while maintaining data utility almost intact, as well as studies to assess the impact of de-identified data on predictive performance. The related work shows a vast number of researches on generalisation and suppression to achieve specific measures for privacy level. In general, experiments have shown that it is possible to protect personal information without the complete degradation of predictive performance. However, the conclusions of such studies are, to a certain extent, limited as many use the same data sets, and although the conclusions in such studies are intuitive, the behaviour of PPT in other types of data sets is not verified. In addition, we notice that only two studies present the consequent re-identification risk after the application of PPT~\citep{vanichayavisalsakul2018evaluation, tania2021compromise}. Besides, just one research does not attempt to reach any specific measure for privacy level~\citep{mivule2013comparative}. We also notice a huge gap in studies on classification with de-identified data using other PPT.

Our approach differs from previous work in several aspects. Such works generally use a specific de-identification algorithm to protect individuals' privacy; in our study, we apply PPT according to data characteristics to evaluate the impact of such techniques individually. Besides that, we de-identify the data using three non-perturbative and two perturbative techniques. To our current knowledge, no study includes as many techniques simultaneously. Furthermore, we analyse the re-identification risk comparing each transformed data set with the original data, as in earlier work~\citep{tania2021compromise}. On top of that, we use a significant amount of data sets as well as learning algorithms. Finally, our main goal is to determine the existence of a trade-off between re-identification ability and predictive performance. 

\section{Data Privacy and Predictive Performance}\label{sec:privperf}

A fundamental problem in privacy-preserving data publishing for learning tasks is identifying the suitable 
privacy-preserving techniques (PPT). In the perspective of privacy loss, the PPT must not allow the original data points to be recovered. On the other hand, from the perspective of utility gain, the predictive performance of the solution must be closely approximated to the model that received the original data set as input.
Finding an optimal balance between privacy and utility could be a challenging issue. In this regard, we formulate the problem that we intend to solve in the following section.

\subsection{Problem Definition}\label{subsec:probdef}

We now provide a formal definition for data privacy, linkability and classification task tackled in this paper. 

Consider a data set $T = \{t_1, ..., t_n\}$, where $t_i$ corresponds to a tuple of attribute values for an individual's record. Let  $V = \{v_1, ..., v_m\}$ be the set of $m$ attributes, in which $t_{i,v_j}$ denotes the value of attribute $v_j$ for tuple $i$. A QI consists of 
attributes values (either categorical or numeric) that could be known to the intruder for a given individual and $QI \in V$. 
An equivalence class corresponds to a combination of records that are indistinguishable from each other, then two tuples, $t_a$ and $t_b$ are in same equivalence class if $t_{a,[QI]} = t_{b,[QI]}$. In the de-identification process, one or more ``transformation functions'' are applied to the original data, producing de-identified data. A data set with minimum privacy guarantees does not have a distinct set of values in a group of attributes for a single individual, i.e., equivalence class of size one, and an intruder cannot isolate any data points or infer private information. 

Although the PPT aim to transform personal information, it may not be enough to protect it from intruders linking private information back to an individual. A link disclosure measure can be described as the amount of correct links that an intruder may infer between original and protected records. With this in mind, the strategy most known used by intruders is record linkage. The linkability by record linkage, proposed by~\citet{fellegi69}, aims to classify the compared pairs of records as matches if two records are supposed to refer to the same entity and non-matches if they refer to a different entity. Assuming two data sets, \textit{A} (original) with \textit{a} elements and \textit{B} (protected) with \textit{b} elements, the comparison space is the product of all possible record pairs as follows. 

\begin{equation}
    A \times B = \{(a, b): a \in A, b \in B\} 
    \label{eq:record_linkage}
\end{equation}

The Equation~\ref{eq:record_linkage} corresponds to the disjoint sets:

\begin{equation}
    M = \{(a, b): a = b, a \in A, b \in B\} 
    \label{eq:m}
\end{equation}

and 

\begin{equation}
    U = \{(a, b): a \neq b, a \in A, b \in B\} 
    \label{eq:u}
\end{equation}

where \textit{M} and \textit{U} correspond to the matched and non-matched sets respectively. 

Several modifications of this approach have been discussed~\citep{nin2007use, christen2009similarity}. We focus on the similarity of record pairs. As such, a similarity function is applied to each pair in resulted product space, where 1 indicates the maximum similarity, which means that two data points are identical. The distance-based record linkage finds, for every protected record $b \in B$, an original record $a \in A$ which minimises the distance to $b$, for the specified similarity function. However, in the case of a larger data set, it is computationally impracticable to consider the entire search space. Therefore, decision rules and a ``blocking phase'' must be defined to reduce into a smaller data set of individuals with at least certain common characteristics.

Suppose a data provider desire to release data to the public for classification tasks. Therefore, in data processing phase, the utility for such an aim should be assessed. The utility for this scenario could be measured through the performance of learning algorithms. The goal of classification tasks is to obtain a good approximation of the unknown function that maps predictor variables toward the target value. The unknown function can be defined as $y=f(x_{1},x_{2},...,x_{p})$, where $y$ is the target variable, $x_1, x_2,...,x_p$ are features and $f()$ is the unknown function we want to approximate. This approximation is obtained using a training data set $D=\{\langle x_{i},y_{i}\rangle\}_{i=1}^n$.

In this paper, we analyse the effectiveness of several PPT in re-identification risk as well the consequent impact on predictive performance. For such an aim, we investigate the existence of a trade-off between the two topics.

\section{Experimental Study}\label{sec:expstudy}

Based on the problem scope detailed in the previous section, we now present our broad experimental study. The main research questions that we intend to answer regarding the defined problem are as follows. 

\begin{enumerate}[start=1,label={\bfseries RQ\arabic*}]
    \item What is the impact of PPT on re-identification ability?
    \item What is the impact of PPT on predictive performance?
    \item Is there a trade-off between re-identification ability and predictive performance?
\end{enumerate}

In the next sections, we present our methodology of the experimental study followed by the presentation of the used data. Afterwards, we present the used methods, including PPT and learning algorithms with respective parameter grids. Then, we present the evaluation of our experiment.

\subsection{Methodology}\label{subsec:methodology}

Our methodology to answer the previous research questions begins with the application of PPT on each original data set. Since we do not know the corresponding category of each attribute in the data sets used in this experiment, we assume all attributes as quasi-identifiers (QI). The PPT are applied according to the characteristics of the QI, i.e., the type of attributes' values. Such a decision avoids tailoring the classification at ``hand'' and depending on the characteristics of the attributes, probably not all of them need to be protected.

In addition, we use a parameter grid to choose the best transformation for a certain PPT according to the level of privacy. From this procedure are created new data sets, each corresponding to a possible combination of PPT that we call transformed variants. As we pinpointed earlier, after the data suffers any transformation, we must evaluate how well the individuals' information is protected. Hence, we assess the re-identification risk in each transformed variant. 

The learning pipeline is then employed to answer the questions on predictive performance. We split into train and test sets each transformed variant using an 80\%/20\% random partition of instances. In the training phase, we use the common estimation k-fold cross validation~\citep{kohavi1995study} that splits the training data into the training and validation subset. This method is used for model assessment in which is established beforehand a grid of hyper-parameters and the k-fold cross validation will select the best parameters. Based on the given result, the model is re-trained with the best parameter. In the end, a final evaluation is performed regarding the predictions on testing data.

The model performance estimation through k-fold cross-validation can be noisy, meaning that each time this method is executed, different splits of the data set into k-folds may occur. Consequently, the distribution of performance scores may vary, resulting in a different mean estimate of model performance. Therefore, we have repeated this process five times to ensure that all original records are included in the testing data set and averaged the results. 

\begin{figure}[!ht]
    \centering
    \includegraphics[width=\linewidth]{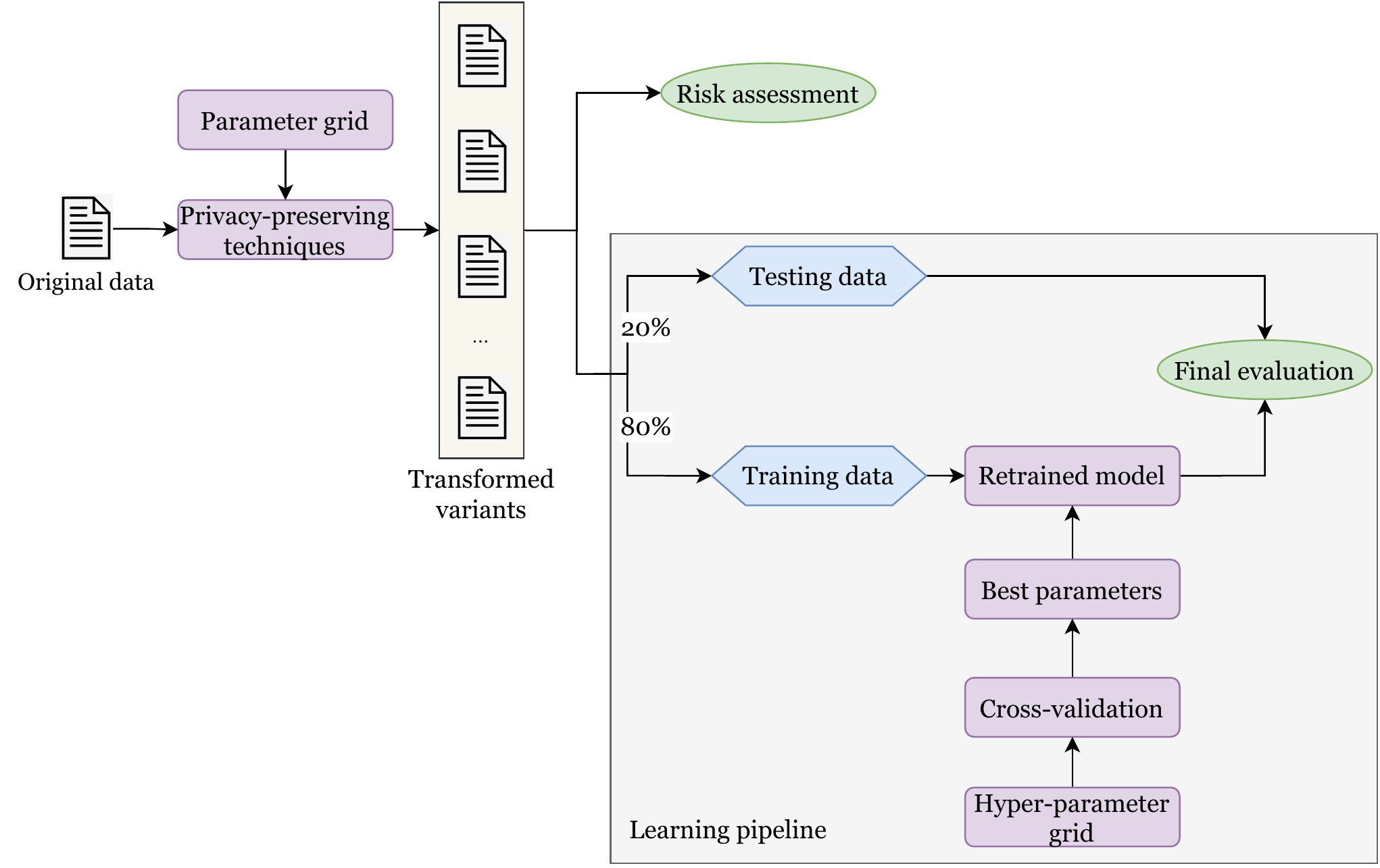}
    \caption{Methodology of the experimental study.}
    \label{fig:methodology}
\end{figure}

To better understand our methodology, Figure~\ref{fig:methodology} illustrates all the steps employed in our experimental study in detail. We highlight that in the learning phase, we removed direct identifiers such as ID or telephone numbers, attributes that have as many distinct values as the number of instances to avoid overfitting.

\subsection{Data}\label{subsec:data}

As we provide an extensive experimental study, we resort to the OpenML~\citep{OpenML2013}, which is a public repository, to collect a diverse group of data. From this platform, we acquired 62 data sets with the following main characteristics.

\begin{itemize}
    \item binary class target;
    \item number of instances between 1000 and 90.000;
    \item number of features greater than 3 and less than 130;
\end{itemize}

In such a set, there is an average of 20 numerical attributes with a standard deviation of 25. Also, there is an average of 4 nominal attributes with a standard deviation of 13. The average size of data sets is 7.572 with a standard deviation of 12.500. One should note that no specific selection was considered. Our goal was to acquire a wide variety of structured data sets for our experiment. Therefore, we collected data related to diverse topics, for example, pathologies identified in the individuals, house characterisation of a given area or even data on how bank customers choose their banks.

\subsection{Methods}\label{subsec:methods}

On the subject of privacy concerns, we select some of the standard PPT to transform the data: suppression, top-and-bottom coding, noise, rounding and global recoding. Such techniques are applied based on the data properties, which are detailed as follows.  

\begin{enumerate}
    \item Suppression was applied in the data sets where the attributes have a certain percentage of distinct values. Although there are no defined criteria to suppress, we created a parameter named \textit{uniq\_per} which specify the limit of the percentage of instances to suppress the whole attribute. For instance, if an attribute has more than 70\% of distinct values, the attribute is suppressed.
    
    \item As some of the data sets present outliers, we deal with these extreme points by applying top-and-bottom codding. In order to detect them, we used the Tukey's method~\citep{tukey1977exploratory}, which defines lower and upper limits from the interquartile range (IQR) and the first ($Q_1$) and third quartiles ($Q_3$). An outlier could be located between the lower and upper inner fence ([$Q_1-(1.5*IQR)$, $Q_3+(1.5*IQR)$]) or between the lower and upper outer fence ([$Q_1-(3*IQR)$, $Q_3+(3*IQR)$]).
    The \textit{outlier} parameter that we created takes the inner and outer fence values to find the outliers. They are then replaced by the minimum and the maximum observations (known as whiskers).
    
    \item The addition of noise was only applied to the floating points. We used the Laplace distribution, which is a well-known distribution for adding random noise~\citep{dwork2008differential, holohan2017k}. For each equivalence class (\textit{ec}) of a floating attribute, we followed the Laplace distribution with mean 0 and variance $diam / \epsilon$. The diameter \textit{diam} of an equivalence class is given by $diam(ec) = \max(ec) - \min(ec)$ and $\epsilon$ corresponds to the privacy budget which determines how strict the privacy is. 
    The \textit{ep} parameter that we created represents this privacy budget.  
    
    \item We also applied the rounding technique. Although rounding is commonly used in integers, we extended its application to the floating points. The \textit{base} parameter specifies the rounding base. 
    
    \item Global re-coding was only applied on integers. We determined the standard deviation for each integer attribute, and next, we constructed ranges by multiplying the standard deviation and a certain magnitude defined by the \textit{std\_magnitude} parameter that we created. To maintain the attributes as an integer, we use the lower limit of the resulted range.
    
\end{enumerate}

We highlight that such techniques were applied in the order they are presented above. Suppose a data set with both nominal and float attributes. If any nominal attribute has a high percentage of unique values, the attribute is suppressed; then, noise is added to float attributes.

Table~\ref{tab:gridTransf} summarises all PPT used in our experimental study and the attribute types in which these were applied, along with the corresponding parameterisation tested. 

\begin{table}[!ht]
\begin{center}
    \begin{adjustbox}{max width=0.85\textwidth}
\begin{tabular}{@{}l|l|l@{}}
\toprule
\textbf{Privacy-preserving techniques} & \textbf{Attribute types} & \textbf{Parameters}          \\ \midrule
Suppression                        & Nominal and numeric                      & uniq\_per: \{0.7, 0.8, 0.9\}    \\ \midrule
Top-and-bottom Coding               & Numeric        & outlier: \{1.5, 3\}          \\ \midrule
Noise                              & Float                    & ep: \{0.5, 2, 4, 8, 16\}     \\ \midrule
Rounding                           & Numeric        & base: \{0.2, 5, 10\}         \\ \midrule
Global re-coding                    & Integer                  & std\_magnitude: \{0.5, 1.5\} \\ \bottomrule
\end{tabular}
\end{adjustbox}  
\caption{Privacy-preserving techniques according the attribute types and respective parameter grid.}
\label{tab:gridTransf}
\end{center}
\end{table}

The PPT parameterisation allows us to perform a series of experiments. To obtain the best parameter, we focus on the impact that each transformation had on individuals' privacy, i.e. the best parameter is the one that presents a higher level of privacy. Such a level was acquired with record linkage that allows us to determine the risk of identity disclosure. This procedure was performed with \textit{Python Record Linkage Toolkit}~\citep{de_bruin_j_2019_3559043} using the default parameter values. Such a tool uses a distance function to verify the similarity between the pairs. The re-identification risk is then obtained by comparing each transformed variant with the original data set and analysing how many records are coincident. The matched records mean that they are at risk. 

When comparing a transformed variant with the original data set, record linkage provides a score ([0, 1]) for each record and each attribute based their similarity. The attributes score is summed up to form a new score for each record. We then select the records with highest score, since the higher the score, the higher the potential matches. Therefore, for the purposes of this work, we assume that the records at highest risk matches at least 70\% of the QI. The best parameter for each PPT is then selected based on the minimum amount of matched records.    

In the learning phase, we select five classification algorithms to test the PPT which includes Random Forest~\citep{ho1998random}, Bagging \citep{breiman1996bagging}, XGBoost~\citep{chen2016xgboost}, Logistic Regression~\citep{logit} and Neural Network~\citep{NN}. We use the \textit{Scikit-learn} python library~\citep{pedregosa2011scikit} to build the classifiers. Final models for each algorithm are chosen based on a 5-fold cross-validation estimation of evaluation scores; therefore, we apply a grid search method with 5*5-fold cross-validation methodology.
Table~\ref{tab:algorithms} details the algorithms and parameter grid used in the experiments.

\begin{table}[!ht]
\begin{center}
    \scriptsize
    \begin{adjustbox}{max width=0.85\textwidth}
\begin{tabular}{@{}l|l@{}}
\toprule
\textbf{Algorithm}  & \textbf{Parameters}                                       \\ \midrule
Random Forest       & \begin{tabular}[c]{@{}l@{}}n\_estimators: \{100, 250, 500\}\\ max\_depth: \{4, 6, 8, 10\}\end{tabular}                                                                                       \\ \midrule
Bagging             & n\_estimators: \{100, 250, 500\}                                                                                            \\ \midrule
Boosting            & \begin{tabular}[c]{@{}l@{}}n\_estimators: \{100, 250, 500\}\\ max\_depth: \{4, 6, 8, 10\}\\ learning\_rate: \{01, 0.01, 0.001\}\end{tabular}                                                                     \\ \midrule
Logistic Regression & \begin{tabular}[c]{@{}l@{}}C: \{0.001, 1, 10000\}\\ max\_iter: \{10000, 1000000\}\end{tabular}                                                                                      \\ \midrule
Neural Network      & \begin{tabular}[c]{@{}l@{}}hidden\_layer\_sizes: \{{[}n\_feat{]}, {[}n\_feat / 2{]}, {[}n\_feat * 2/3{]},\\                                      {[}n\_feat, n\_feat / 2{]}, {[}n\_feat, n\_feat * 2/3{]}, {[}n\_feat / 2, n\_feat *  2/3{]},\\                                      {[}n\_feat, n\_feat  / 2, n\_feat  * 2/3{]}\}\\ alpha: \{0.05, 0.001,  0.0001\}\\ max\_iter: \{10000, 1000000\}\end{tabular} \\ \bottomrule
\end{tabular}
\end{adjustbox}
\end{center}    
\caption{Learning algorithms considered in the experimental study and respective hyper-parameter grid. Each line of parameter \textit{hidden layer sizes} (Neural Networks) represents an increasing number of layers (1--3).}
    \label{tab:algorithms}
\end{table}

\subsection{Evaluation}\label{subsec:eval}

Multiple evaluation metrics have been proposed to evaluate the performance in classification tasks. These include Accuracy, F-score~\citep{Rijsbergen_fmeasure}, geometric mean~\citep{kubat1998machine}, AUC (Area Under the ROC Curve)~\citep{weng2008new}, among many others. Given the evaluation metrics available and the dimension of our experimental study, we focus on using the F-score metric.  

Furthermore, we apply statistical tests using Bayes Sign Test~\citep{Benavoli2014,Benavoli2017} to evaluate the significance of the F-score results. For such a goal, we use the percentage difference between each pair of solutions as follows:

\begin{equation}
    \frac{R_a - R_b}{R_b} * 100
\end{equation}

where, $R_a$ is the solution under comparison and $R_b$ is the baseline solution.

The ROPE (Region of Practical Equivalence)~\citep{kruschke2015bayesian} is then used to specifying an interval where the percentage difference is considered equivalent to the null value. We apply the interval [-1\%, 1\%] for ROPE, as such interval is the common used in classification tasks. 
Therefore, \textit{i)} if the percentage difference of F-score between solutions a and b (baseline) is greater than 1\%, the solution \textit{a} outperforms the \textit{b} (win); \textit{ii)} if the percentage difference is within the specified interval, they are of practical equivalence (draw); and \textit{iii)} if the percentage difference is less than -1\%, the baseline outperforms solution \textit{a} (lose).

\subsection{Results: Re-identification Risk}\label{subsec:resA}

From 62 original data sets, we obtained 1074 transformed variants. In this set, all possible combinations of the five PPT previously presented are considered. Accordingly, each original data set can produce between 7 and 31 transformed variants depending on their characteristics.

\begin{figure*}[!ht]
   \centering
   \includegraphics[width=0.7\linewidth]{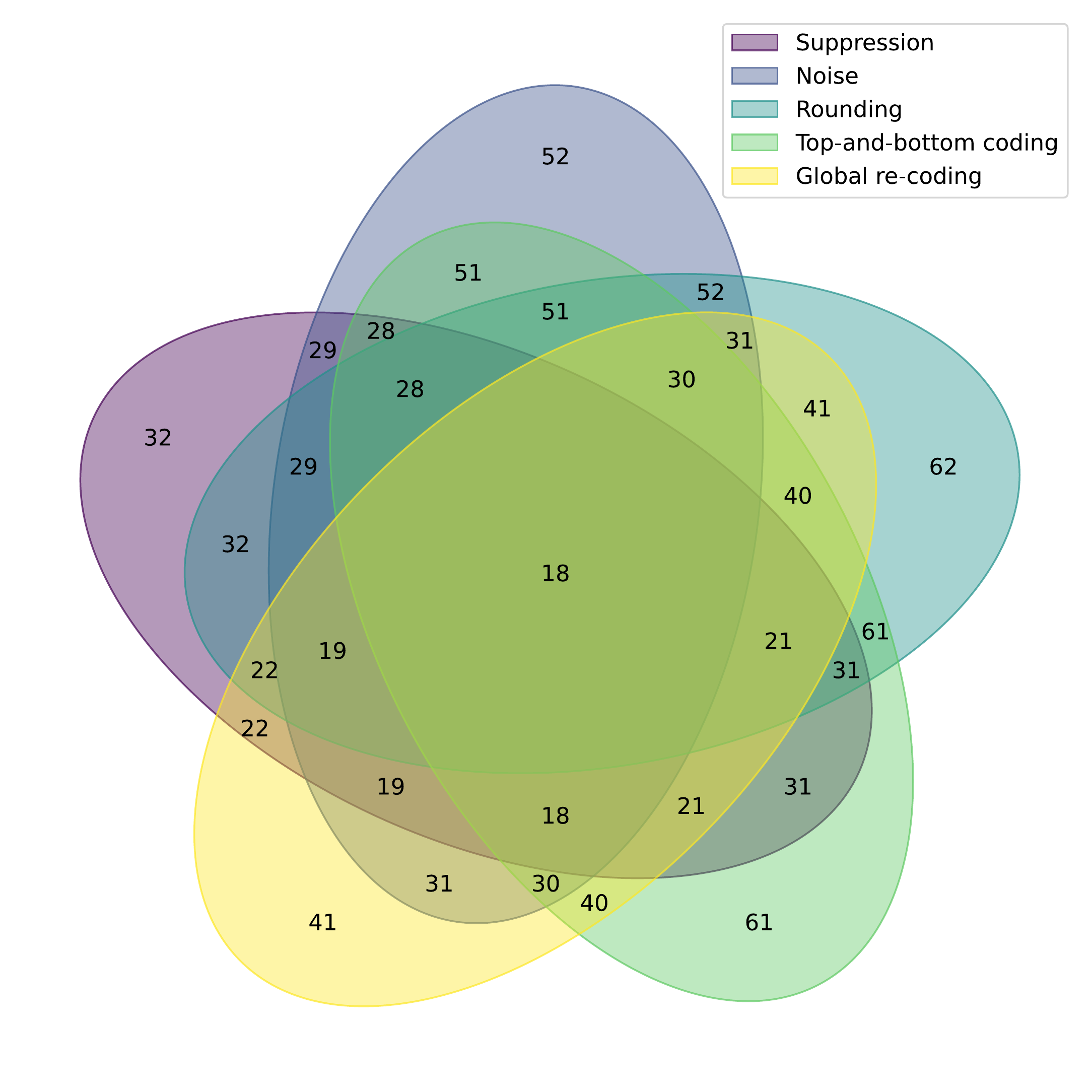}
 \caption{Venn diagram on all data sets generated from the application of the privacy-preserving techniques.}
 \label{fig:venn}
\end{figure*}

Figure~\ref{fig:venn} displays the number of transformed variants with the corresponding PPT. As all original data sets present numerical features (integer and floating points), we applied rounding to all of them. On the other hand, top-and-bottom coding has only not been tested to one data set, and the same happens in combining this technique with rounding. The least present technique was suppression; just over half of original data sets have attributes with a high percentage of distinct values. The combination of global re-coding and suppression is only applied to 21 data sets. We highlight that of the 62, only 18 have the necessary characteristics to be transformed with the five techniques simultaneously. Such a lower number of data sets with all possible solutions is due to the suppression applied to so few data sets. 

On the one hand, there are original data sets in which not all PPT have been applied, and on the other hand, we have 18 original data sets in which each resulted in 31 transformed variants. Henceforth, we focus our comparisons on the results across the 18 data sets and on all data sets that include a specific PPT. For better visualisation of results, we abbreviate the name of the PPT to the following nomenclature: G (Global re-coding), N (Noise), R (Rounding), S (Suppression) and T (top-and-bottom).

\begin{figure*}[!ht]
   \centering
   \scriptsize
   \includegraphics[width=\textwidth]{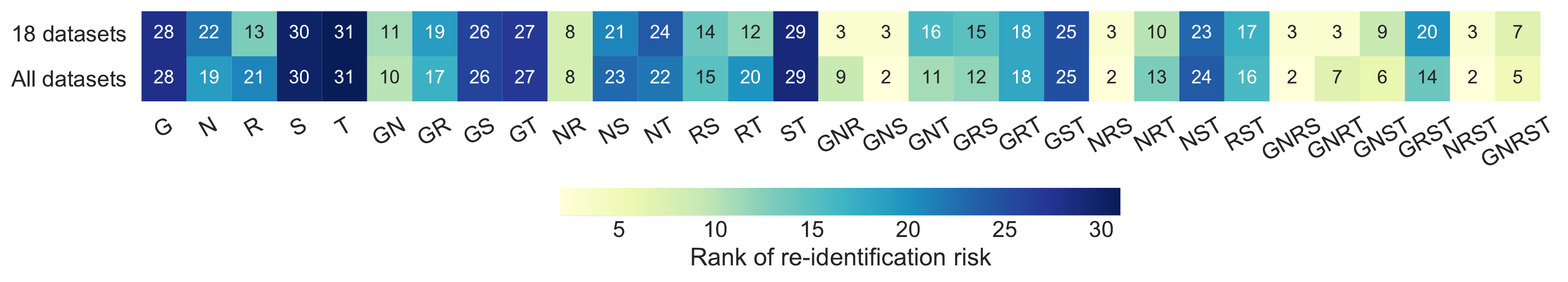}
 \caption{Rank of re-identification risk for each transformed variant.}
 \label{fig:risk}
\end{figure*}

The effectiveness of the previous techniques applied is measured through the re-identification risk. Figure~\ref{fig:risk} presents the rank value of the averaged re-identification risk for each transformed variant. A high rank means that the respective technique has the highest average re-identification risk. Therefore, top-and-bottom present the highest risk. This result is expected since only the outliers of each data set are modified. Data sets with suppression also present a high re-identification risk because many of these data sets also have attributes of floating point type. Such attributes make most of the records unique. 

Regarding single PPT, rounding and noise are the least risky. This result is due to the transformation of floating points, making these attributes less granular. Under those circumstances, variants with the combination of rounding and noise present the highest level of privacy. 

In the perspective of electing the best PPT, we notice in Figure~\ref{fig:risk} that there are six solutions with the lowest rank regarding the 18 data sets, which means that there are six possible solutions to guarantee privacy of the individuals. However, each one may have a different impact on utility. Therefore, we will analyse in the following section the results concerning the utility evaluation.


\subsection{Results: Predictive Performance}\label{subsec:resB}

\begin{figure*}[!ht]
   \centering
   \includegraphics[width=\textwidth]{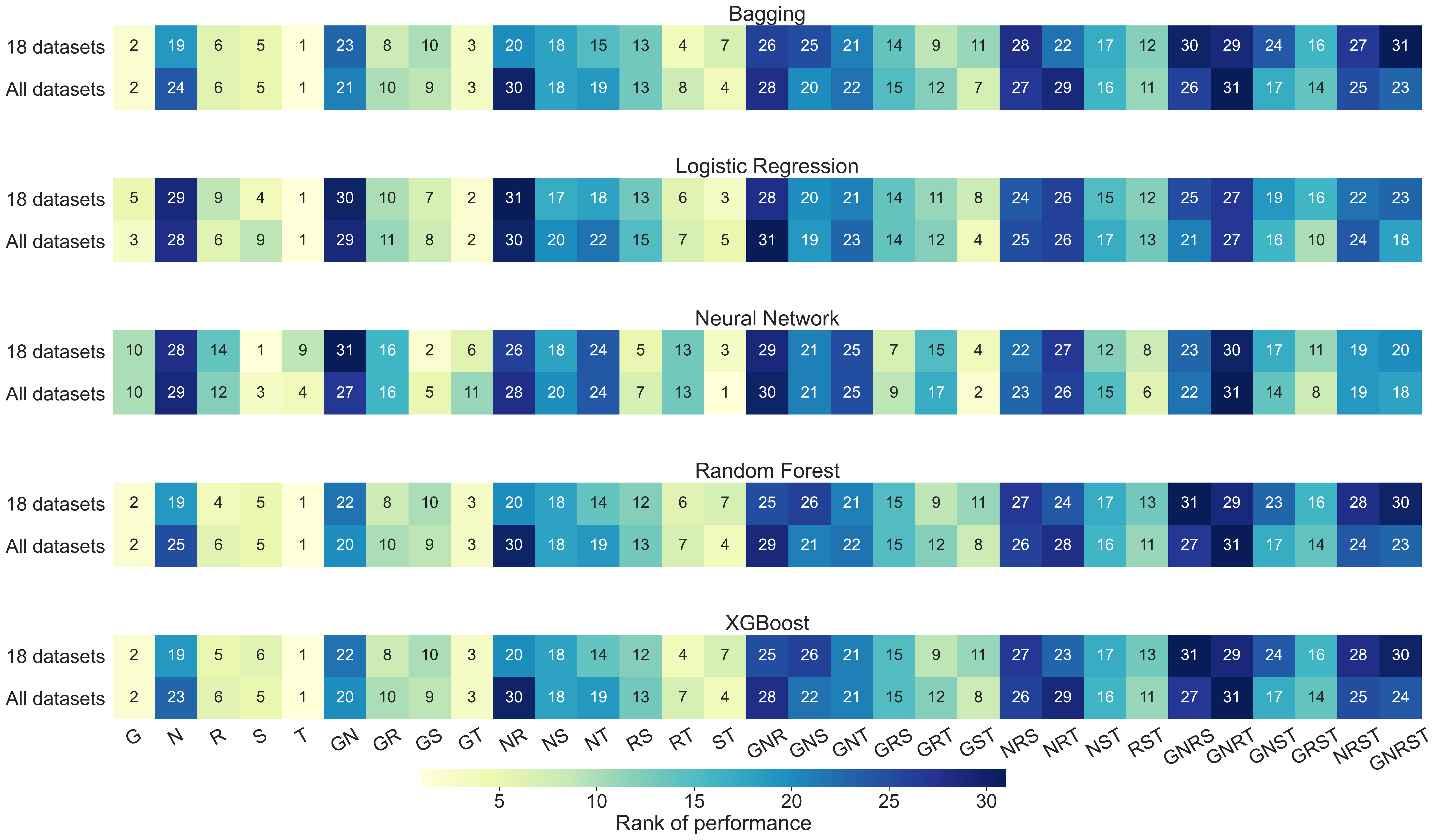}
 \caption{Rank of predictive performance for each transformed variant.}
 \label{fig:performance}
\end{figure*}

The results of the learning algorithms are presented regarding the F-score metric.
Hence, we selected the averaged F-score in the validation set for each transformed variant, and for each PPT, we averaged the previous F-score and then obtained the rank. Figure~\ref{fig:performance} displays such results also considering all data sets against the 18 data sets, where a lower rank corresponds to the best predictive performance.

In contrast to the re-identification risk results, we can verify that top-and-bottom is the technique that shows the best performance in a general perspective. This outcome is due to the slight change (outlier transformation) maintaining the similarity to the original data. However, with Neural Networks, this result is not verified; it is the suppression technique that demonstrates the best predictive performance. In suppression, attributes with a high percentage of unique values are removed. Thus, data sets without these types of attributes may have a more accurate prediction. A possible explanation for such results is the feature selection that neural networks perform with suppression. 

We noticed a big difference between noise and the other four techniques when comparing the single PPT. 
Noisy data may deteriorate the predictive performance depending on the degree of sensitiveness to data perturbations of the learning algorithm. For example, neural networks are known to be sensitive to data perturbations~\citep{laugros2020addressing}.  

Comparing the behaviour between the learning algorithms, Bagging, Random Forest, and XGBoost, they behave similarly. On the other hand, Logistic Regression and Neural Network are more sensitive to perturbed data. But in general, when comparing all data sets to the 18 data sets, we note that each algorithm behaviour did not vary by much.

Overall, we verify that the PPT that does not present predictive performance degradation is the one that presents the slightest changes in the data set. Consequently, the combination of techniques results in a greater negative impact on predictive performance. In short, the more transformation in the data the greater the loss of predictive performance.

\subsection{Results: Experimental Settings}\label{subsec:resC}
Next, we analyse our methodology concerning validation and oracle settings. We stress that both experimental settings were applied to the 18 data sets that include all possible combinations of PPT. In validation setting, the objective is to evaluate the solutions generated from the 5-fold cross-validation w.r.t the validation set by estimating the best possible hyper-parameter configurations for each data set. Based on the validation results, we carry out the assessment concerning unseen data (test set). 
On the other hand, the oracle setting objective is to assess the best possible hyper-parameter configurations in an out-of-sample context. Therefore, the original data sets are split into train and test sets and each hyper-parameter configuration is used to create a predictive solution with the training data and evaluated in the test set. The best possible solution in each data set is the \textit{oracle}. For each setting, our focus towards evaluating the best solution based on three main aspects: 
\begin{enumerate}
    \item Predictive performance - for each transformed variant, we calculated the percentage difference between each solution (candidate) and the best solution of the original data set (baseline); 
    \item Transformed variant - the baseline, in this case, corresponds to the best solution of each transformed variant. The percentage difference was obtained between each candidate and the selected baseline;
    \item Re-identification risk - for each original data set, we selected the correspondent transformed variant that guarantee a high level of data protection. After, we selected the highest F-score value for these transformed variants with lower risk. Such a result correspond the baseline used to calculate the percentage difference.
\end{enumerate}

\begin{figure*}[!ht]
   \centering
   \includegraphics[width=\textwidth]{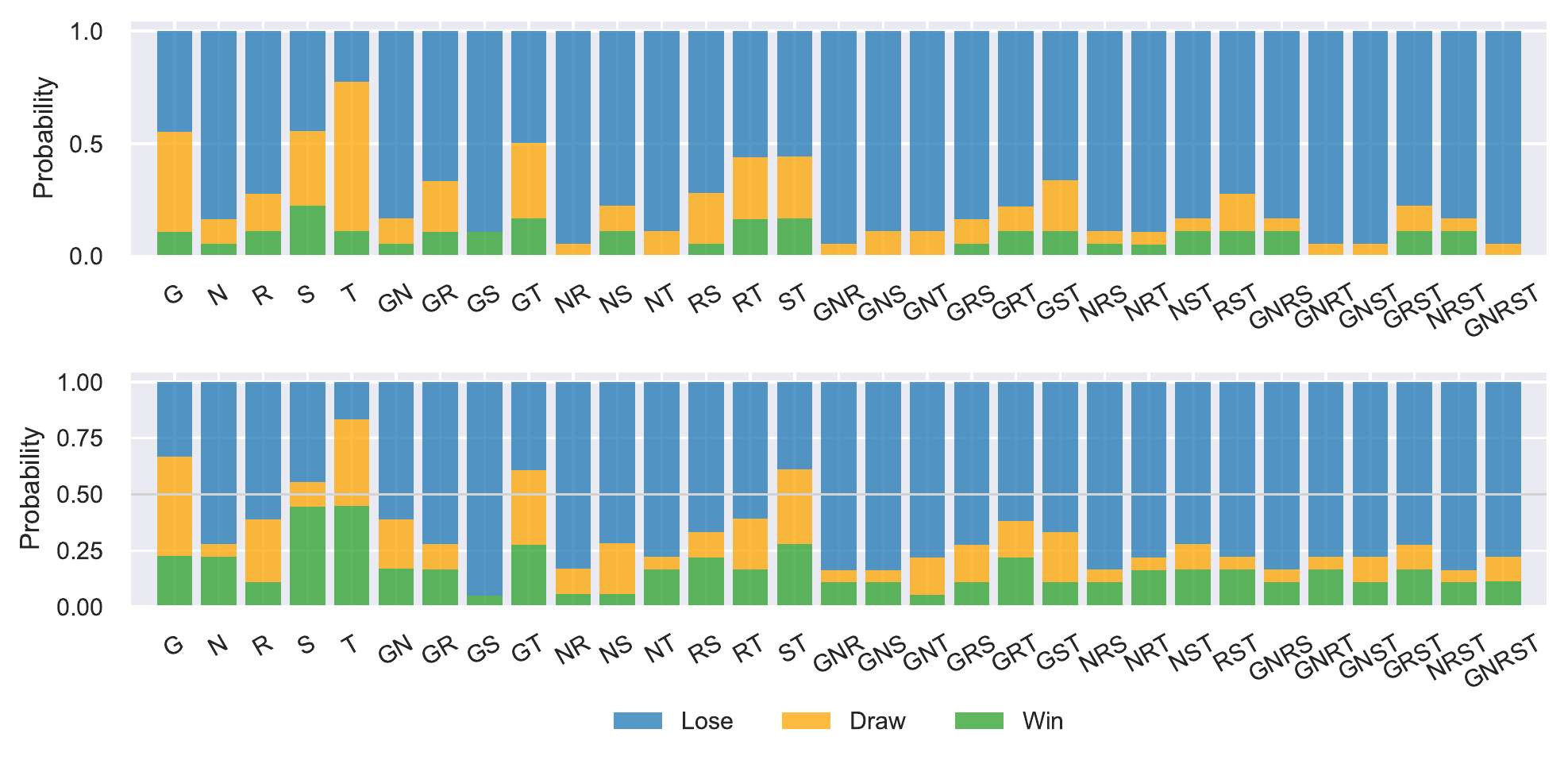}
 \caption{Comparison of each privacy-preserving technique based on results from validation and oracle setting, respectively. Illustrates the proportion of probability for each candidate solution winning, drawing, or losing significantly against the best solution in original data sets, according to the Bayes Sign Test.}
 \label{fig:orig}
\end{figure*}

Our results were analysed using the Bayes Sign Test~\citep{Benavoli2014,Benavoli2017}, with a ROPE interval of [-1\%, 1\%] (details in Section~\ref{subsec:eval}). Figure~\ref{fig:orig} illustrates the outcome of best solution for validation and oracle settings, respectively. In this scenario, the baseline is the original data sets. In a general overview, results demonstrate that baseline solutions outperform the transformed variants. Nevertheless, top-and-bottom coding, suppression, and global re-coding show an advantage compared to other PPT. Such results are supported by Figure~\ref{fig:performance} where these techniques presented the best rank. 
In the oracle setting, we can verify that also global re-coding and top-and-bottom combined, and suppression with top-and-bottom are either of practical equivalence to or outperform the original results with a probability greater than 50\%.    

\begin{figure*}[!ht]
   \centering
   \includegraphics[width=\textwidth]{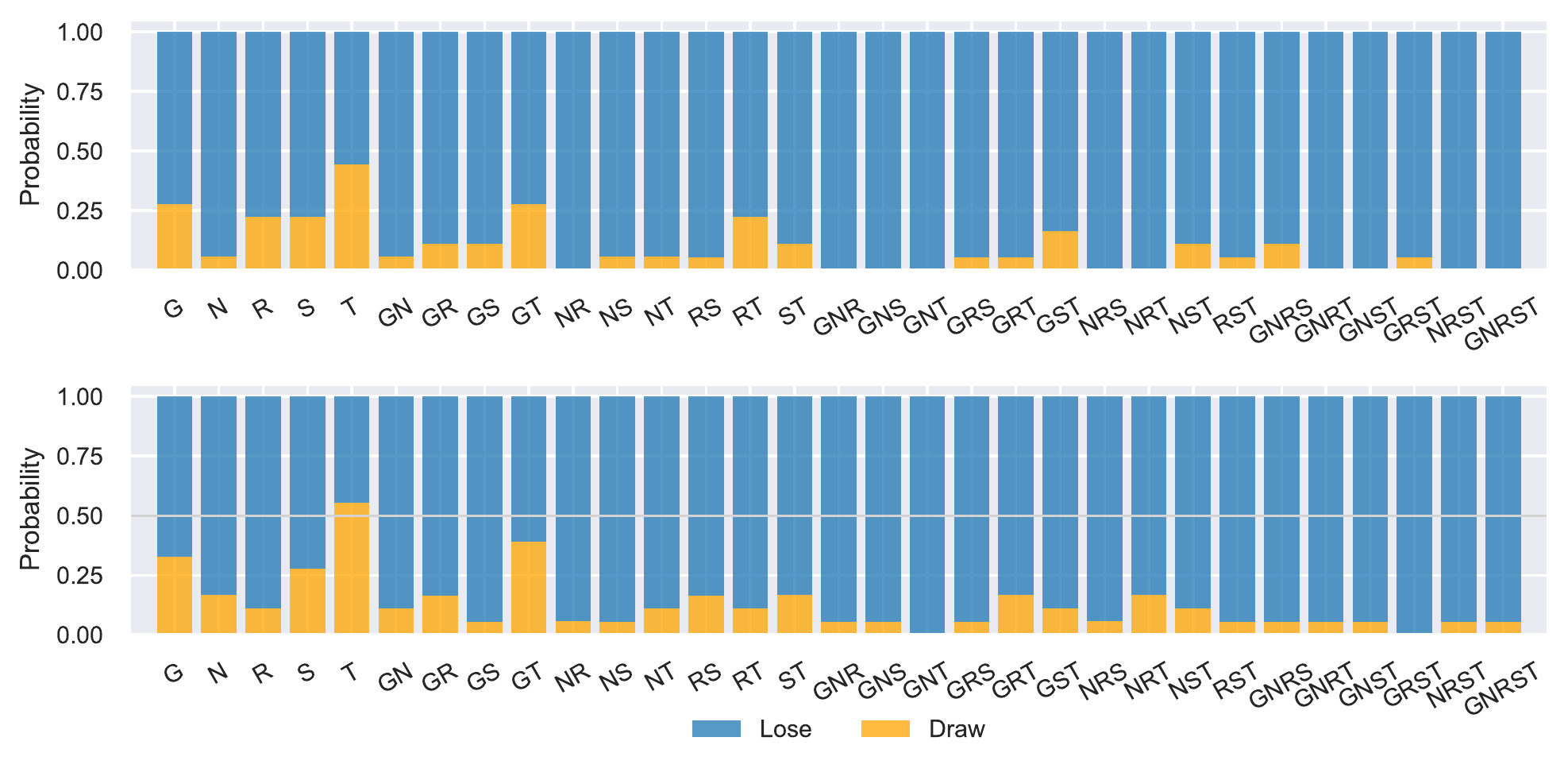}
 \caption{Comparison of each privacy-preserving technique based on results from validation and oracle setting, respectively. Illustrates the proportion of probability for each candidate solution drawing or losing significantly against the best technique, according to the Bayes Sign Test.}
 \label{fig:transf}
\end{figure*}

The following results illustrated in Figure~\ref{fig:transf}, concern the best solution where the baseline is the best solution for each transformed variant in both settings. Such results show that baseline results outperform all transformed variants. Consequently, there are some transformed variants with no equivalent results to the baseline in both scenarios. Also, we notice a slight advantage of oracle over the validation setting. Given that the comparison is made w.r.t the best-transformed variant solution for validation and oracle settings, we do not account for “win” scenarios.

\begin{figure*}[!ht]
   \centering
   \includegraphics[width=\textwidth]{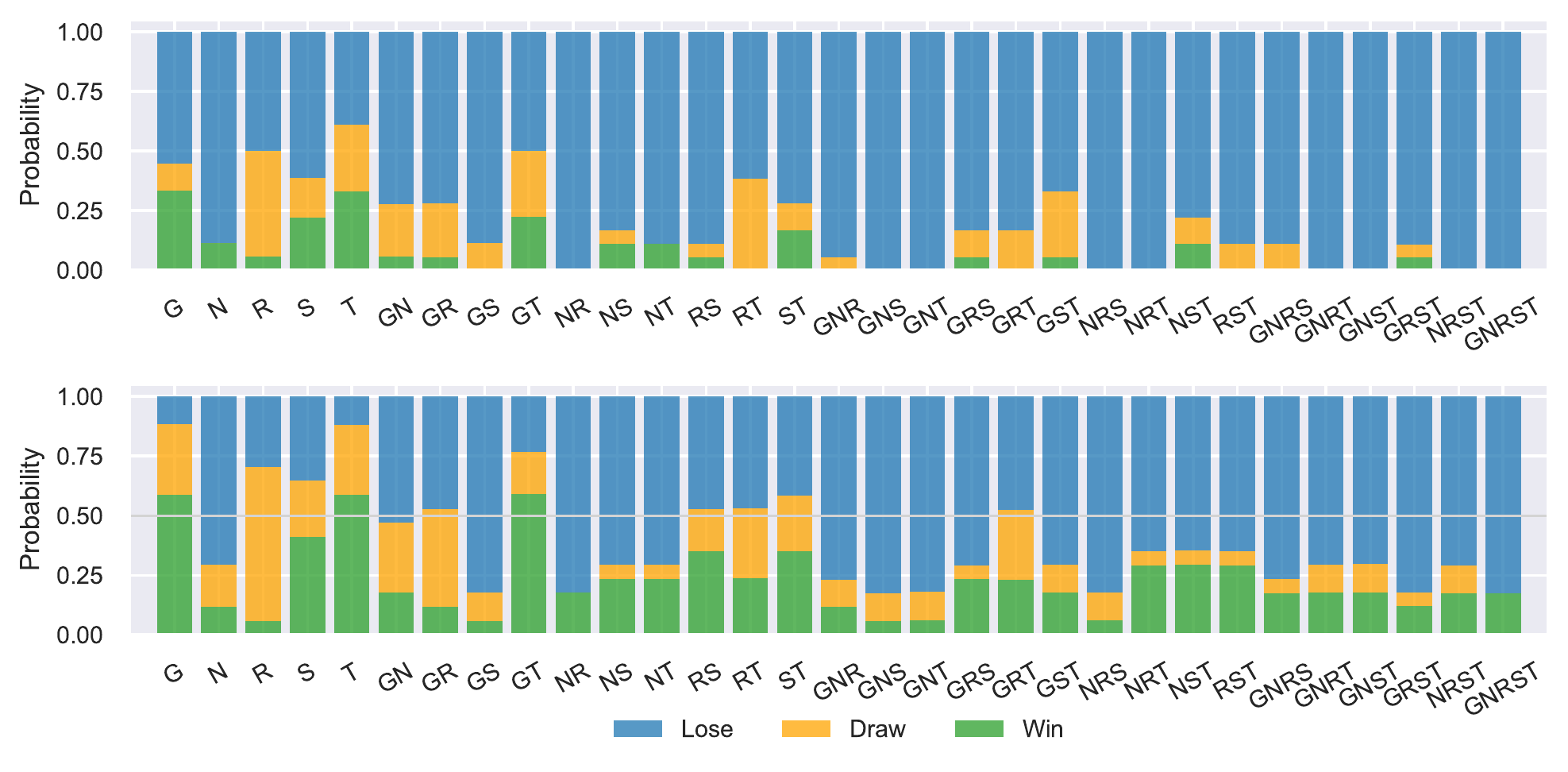}
 \caption{Comparison of each privacy-preserving technique based on results from validation and oracle setting, respectively. Illustrates the proportion of probability for each candidate solution winning, drawing, or losing significantly against the highest protected solution, according to the Bayes Sign Test.}
 \label{fig:priv}
\end{figure*}

The last scenario shows the comparison between all solutions with the best solution at the level of privacy -- lowest re-identification risk. In the case of equal re-identification risk values, the tie-breaker is based on the highest predictive performance. The results are illustrated in Figure~\ref{fig:priv}. We can verify that such results are, in general, worse than the first scenario since there are some transformed variants with no equivalent results to the baseline. There is also a clear advantage of oracle over validation setting in this scenario, especially with global re-coding, top-and-bottom, and the combination of both techniques presenting a probability of winning more than 50\%.

\section{Discussion}\label{sec:discussion}

We propose to answer three research questions concerning the protection of personal information and data utility. In such regard, we analysed the ability on re-identification after the transformation of private information (RQ1). The results showed that a single PPT is not enough to protect the data; its combination presented a lower re-identification risk. However, we highlight that combinations of larger sizes do not mean more privacy; we noticed that combining the five PPT presented a higher re-identification risk than other combinations of smaller sizes. Furthermore, we observed that combinations with both global re-coding and noise, or noise and rounding, are strong candidates to achieve the desired level of privacy.

Succeeding the data transformation to protect the subject's privacy, we proceeded with the predictive performance analysis for all transformed variants to evaluate the cost on utility (RQ2). In our experimental results, we observed that single PPT, except noise, presented a better outcome. We also notice that combinations with top-and-bottom presented a better rank of predictive performance due to the slight modifications of original data. 
In a general perspective, we noticed that predictive performance results were the opposite of re-identification risk.  

After the re-identification risk and predictive performance analysis for each transformed variant, we conclude that there are combinations of PPT that guarantee the desired privacy. However, the same combinations presented a negative impact on predictive performance when compared with the original data sets. Such results mean that the more privacy we guarantee, the more reduced the predictive performance. To support our conclusions, we used a significance test that allows us to conclude that PPT significantly negatively impact predictive performance. Therefore, we conclude that there is a trade-off between re-identification ability and predictive performance (RQ3).  

\begin{figure}[ht!]
   \centering
   \includegraphics[width=0.8\textwidth]{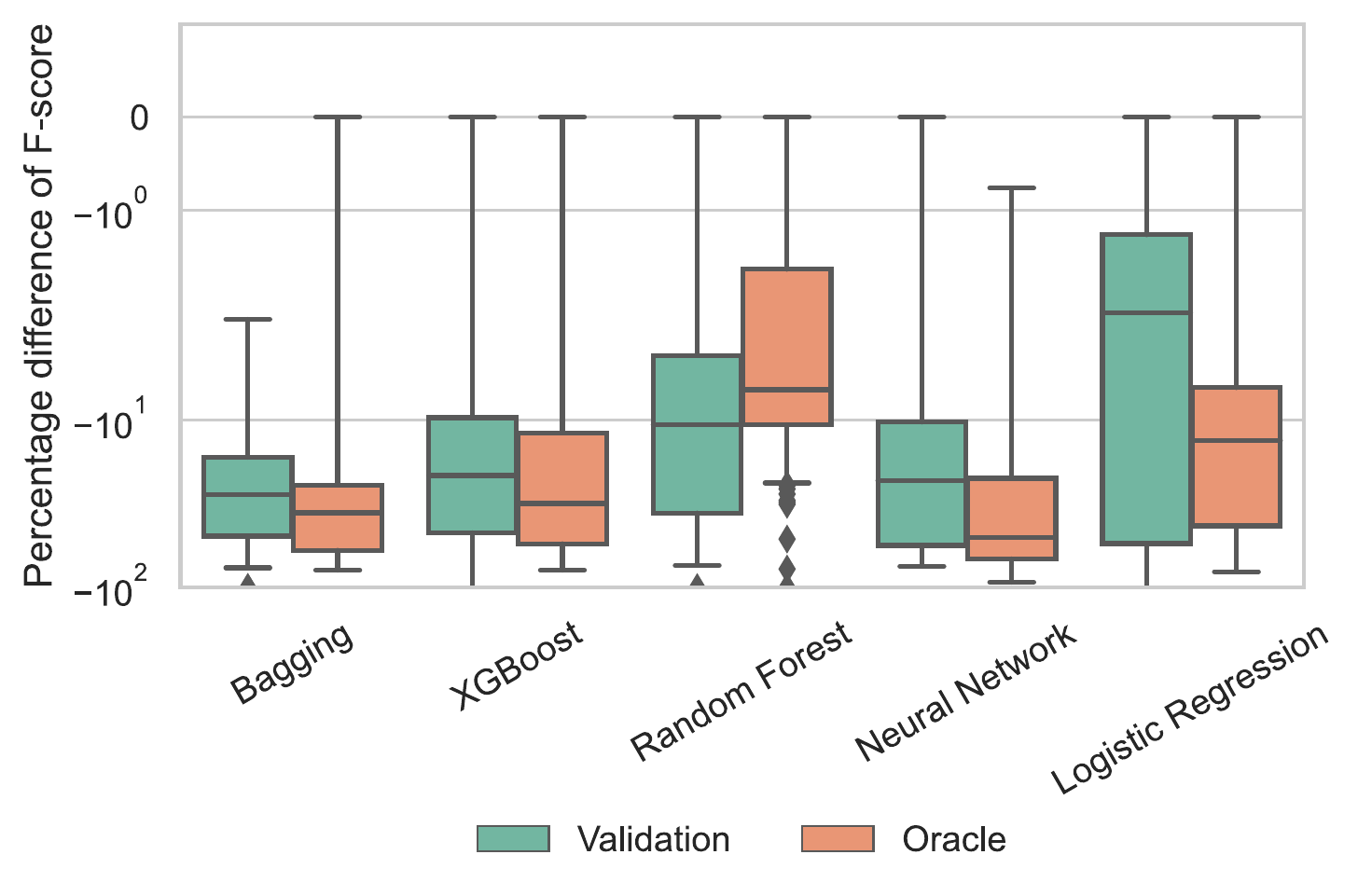}
 \caption{Percentage difference of F-score between each transformed variant with a lower re-identification risk and the original data sets.}
 \label{fig:boxplot}
\end{figure}

Regarding such a conclusion, our aim at this stage is to explore the results for which privacy is maximum, as well as the predictive performance for the 18 data sets in validation and oracle settings. Hence, we selected the transformed variants with lower re-identification risk, and for each, we choose the solutions with the best F-score. Figure~\ref{fig:boxplot} presents the distribution of percentage difference for each learning algorithm. We observe that the median is at a high negative percentage difference level (-100\%) for almost all algorithms tested. In general, we conclude that PPT presenting a lower re-identification risk also present a lower predictive performance. 

From a general perspective, our conclusions show that  the application of combined PPT provides better protection but in the flip side results in a negative impact on predictive performance.
We highlight that such results could have implications for future decisions. For instance, organisations or research teams will have to choose which criterion they consider more relevant -- if it is more critical to guarantee privacy, they may lose predictive performance. However, there is the possibility of assigning two thresholds, one to the risk of identity disclosure and another to the predictive performance, being aware of the risk they run regarding the possible loss of privacy or degradation of data utility. 

Although it is not within our scope, such a conclusion leads us to the topic of information theory~\citep{shannon1948mathematical}, a mathematical representation of the factors affecting the processing of information. Some papers relate the application of PPT to information loss. For instance,~\citet{de1999information} discuss the information loss caused by global re-coding and suppression concerning Entropy. Also,~\citet{gionis2008k} introduce a variation of entropy, called Non-Uniform Entropy, which respects monotonicity. This measure compares frequencies of attribute values in the protected data set with the frequencies of original values. 

\section{Conclusions}\label{sec:conclusions}

To achieve reliable levels on predictive performance, the machine learning models often require huge data sets with detailed information that may contain personal and sensitive information. In such a case, data should be adequately protected with privacy-preserving techniques (PPT) in order to comply with current data protection regulations and consequently reduce risks of personal data breaches. 

In this study, we created a wide range of transformed variants from original data sets using three non-perturbative and two perturbative PPT. Contrary to previous work, such techniques were applied based on data characteristics instead of a de-identification algorithm which needs attributes classification beforehand. Furthermore, we evaluate the impact of each possible combination of PPT in re-identification risk and predictive performance. Our main goal was to evaluate the existence of a trade-off between the re-identification ability and predictive performance. We demonstrated that it is possible to guarantee a null risk of identity disclosure regarding the sample of each data set. We also observe combinations of PPT that guarantee equivalent or even better results than the original data sets due to the slight changes on the original data set or the suppression of attributes with many distinct values. However, our experimental results indicate that the lower the re-identification risk, the lower the predictive performance, and vice-versa. Such results mean that there is evidence of a trade-off between the ability of re-identification and predictive performance. Future challenges include an analysis of this phenomenon from the perspective of information theory and of privacy measures for other types of disclosure, namely, attribute disclosure. 

\section*{Acknowledgments}
The work of Tânia Carvalho is supported by Project ``POCI-01-0247-FEDER-041435 (Safe Cities)'' and financed by the COMPETE 2020, under the PORTUGAL 2020 Partnership Agreement, and through the European Development Fund (EDF). The work of Nuno Moniz is financed by National Funds through the Portuguese funding agency, FCT - Fundação para a Ciência e a Tecnologia, within project UIDB/50014/2020. The work of Luís Antunes is supported by EU H2020-SU-ICT-03-2018 Project No. 830929 CyberSec4Europe (cybersec4europe.eu). Pedro Faria is supported by Project ``DataInvalue - NORTE-01-0247-FEDER-069497''.

\bibliographystyle{model5-names}
\biboptions{authoryear}
\bibliography{bibtex}

\end{document}